\documentclass[10pt,twocolumn,letterpaper]{article}
\pdfoutput=1

\usepackage[pagenumbers]{cvpr} 

\usepackage[utf8]{inputenc} 
\usepackage[T1]{fontenc}    
\usepackage{graphicx}
\usepackage{amsmath}
\usepackage{amssymb}
\usepackage{mathtools}
\usepackage{booktabs}
\usepackage{caption}
\usepackage{subcaption}
\usepackage{float}
\usepackage{enumitem}
\usepackage{bbm}
\usepackage[percent]{overpic}
\usepackage[dvipsnames]{xcolor}

\newcommand{\bx}{\mathbf{x}}
\newcommand{\bz}{\mathbf{z}}
\newcommand{\bu}{\mathbf{u}}
\newcommand{\bX}{\mathbf{X}}
\newcommand{\Teta}[0]{\mathbf{\Theta}}

\DeclareMathOperator*{\argmin}{argmin}
\newcommand{\parag}[1]{\paragraph{#1}}

\newif\ifdraft
\draftfalse
\drafttrue 

\ifdraft
 \newcommand{\PF}[1]{{\color{red}{\bf PF: #1}}}
 
 \newcommand{\RL}[1]{{\color{blue}{\bf RL: #1}}}
 
 \newcommand{\BG}[1]{{\color{olive}{\bf BG: #1}}}
 
 \newcommand{\CD}[1]{{\color{Peach}{\bf CD: #1}}}

 \newcommand{\TODO}[1]{\textbf{\color{red}[TODO: #1]}}
\else
 \newcommand{\PF}[1]{}
 
 \newcommand{\RL}[1]{}
 
 \newcommand{\BG}[1]{}
 
 \newcommand{\CD}[1]{}
 
 \newcommand{\TODO}[1]{}
 
\fi

\definecolor{cvprblue}{rgb}{0.21,0.49,0.74}
\usepackage[pagebackref,breaklinks,colorlinks,citecolor=cvprblue]{hyperref}
\usepackage[capitalize]{cleveref}
\Crefname{equation}{Eq.}{Eqs.}
\Crefname{figure}{Fig.}{Figs.}
\Crefname{table}{Tab.}{Tabs.}
\Crefname{section}{Sec.}{Secs.}

\def\paperID{6094} 
\def\confName{CVPR}
\def\confYear{2024}

\title{Garment Recovery with Shape and Deformation Priors}

\author{
  Ren Li
  \quad
  Corentin Dumery
  \quad
  Benoît Guillard
  \quad
  Pascal Fua\\
  {CVLab, EPFL} \\ 
  {\small \texttt{\{name.surname\}@epfl.ch}}
}

\begin{document}


\twocolumn[{
\renewcommand\twocolumn[1][]{#1}
\maketitle
\begin{center}
    \vspace{-5mm}
    \begin{tabular}{c}
        \begin{overpic}[width=0.99\textwidth]{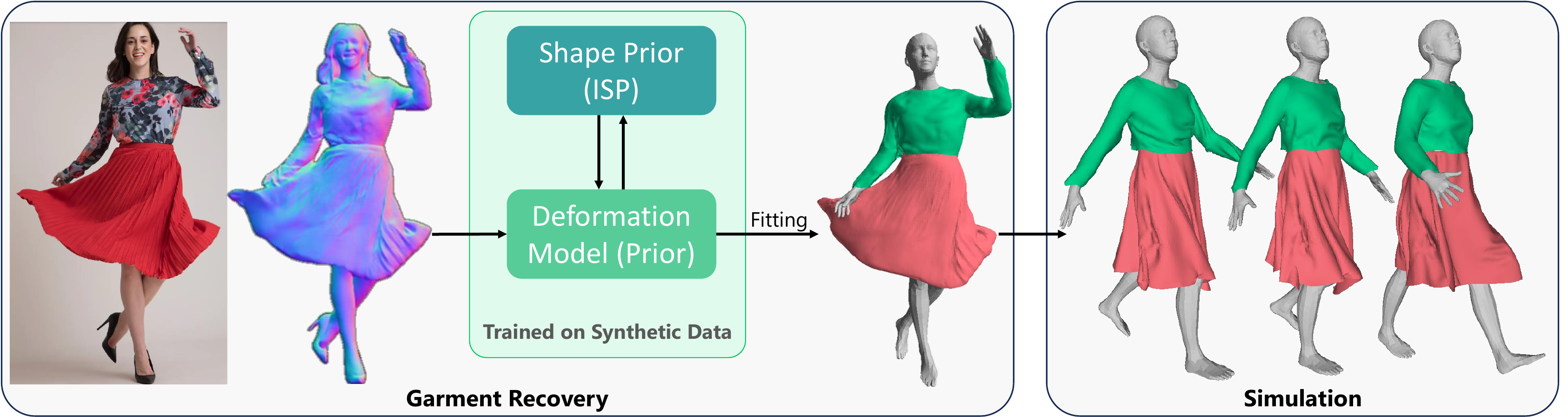}
        \end{overpic} \\
    \end{tabular}
\end{center}
\vspace{-3mm}
\small \hypertarget{fig:abstract}{Figure 1. We propose a fitting method that leverages shape and deformation priors trained on synthetic data to recover realistic 3D garment mesh from in-the-wild images. We produce triangulated meshes that are directly usable for animation and simulation.}
\vspace{0.3cm}
}]

\begin{abstract}

While modeling people wearing tight-fitting clothing has made great strides in recent years, loose-fitting clothing remains a challenge. We propose a method that delivers realistic garment models from real-world images, regardless of garment shape or deformation. To this end, we introduce a fitting approach that utilizes shape and deformation priors learned from synthetic data to accurately capture garment shapes and deformations, including large ones. Not only does our approach recover the garment geometry accurately, it also yields models that can be directly used by downstream applications such as animation and simulation.

\end{abstract}


\section{Introduction}
\label{sec:intro}

3D clothed human recovery aims to reconstruct the body shape, pose, and clothing of many different people from images. It is key to applications such as fashion design, virtual try-on, 3D avatars, along with virtual and augmented reality. Recent years have seen tremendous progress in modeling people wearing tight-fitting clothing both in terms of body poses~\cite{Bogo16,Lassner17a,Omran18,Kanazawa18a,Kolotouros19,Georgakis20,Moon20,Joo20,Yang20,Li21i,Li21g} and 3D shape of the clothes~\cite{Danerek17,Bhatnagar19,Jiang20d,Corona21,Moon22,Li22c,DeLuigi23,Li23a}. However, loose-fitting clothing remains a challenge. Existing approaches either rely on mesh templates with limited generality, or produce models expressed in terms of 3D point clouds~\cite{Ma22a} or as a single watertight mesh that tightly binds the body and garment together~\cite{Xiu23}, neither of which is straightforward to integrate into downstream applications.

In this paper, we propose a method that overcomes these limitations and can effectively recover the shape of loose fitting garments from single images. The recovered garments can then be animated without any additional processing, as shown in Fig. \hyperlink{fig:abstract}{1}. Starting from the Implicit Sewing Patterns (ISP) model~\cite{Li23a} that represents garments in terms of  a set of individual 2D panels and 3D surfaces associated to these panels, we introduce a deformation model that we apply to the 3D surfaces so that they can deviate substantially from the body shape. These deformations are conditioned on normals estimated from an input image of the target garment. They are learned 
from synthetic mesh data featuring loose clothing, where the deformations are taken to be those required to fit individual ISP 3D surfaces to the ground-truth 3D meshes.

Given the trained deformation model, we designed a two-stage fitting process to recover the 3D garment from in-the-wild images. First, the parameters of the pre-trained deformation model are optimized to produce a shape that minimizes the distance between garment outlines and segmented garment regions,
the differences between garment normals and the normals estimated in the images by off-the-shelf-algorithms~\cite{Xiu23,Saito20a}, and a physics-based loss to promote physical plausibility of the results. Then, fine local details are recovered by directly optimizing the vertex positions of the reconstructed mesh with the same loss. Our fitting process does not require external 3D annotations, other than the estimated normals of the target garment.

We demonstrate that our method can recover garments that go from tight- to loose-fitting and outperforms existing approaches~\cite{Jiang20d,Corona21,Moon22,DeLuigi23,Li23a,Xiu23} in terms of reconstruction accuracy. Furthermore, our reconstructed meshes are directly usable for virtual try-on or animation, unlike \cite{Ma22a,Xiu23}. Our implementation and model weights are available at \url{https://github.com/liren2515/GarmentRecovery}.

\begin{figure*}[h]
    \centering
    \includegraphics[width=0.98\textwidth]{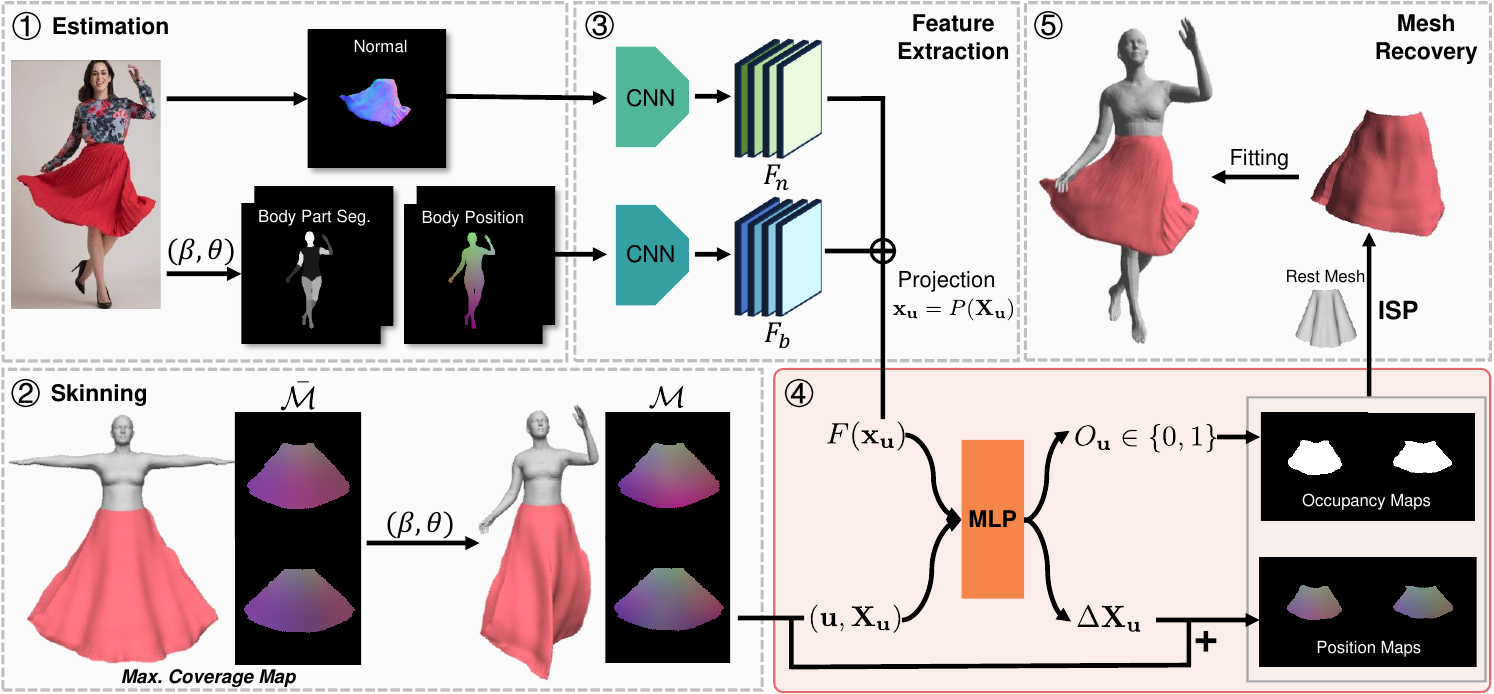}
    \vspace{-2mm}
    \caption{\textbf{Framework.} Given an image, (1) we first estimate the normal map of the target garment and the SMPL body parameters $(\beta,\theta)$, which are used to compute the body part segmentation and position maps. (2) The maximum coverage garment shape $\bar{\mathcal{M}}$ is then skinned to closely fit to the body, yielding $\mathcal{M}$. Leveraging (3) pixel-aligned image features, our deformation model (4) predicts occupancy and position maps to correct $\mathcal{M}$ for large deformations. (5) The 3D garment mesh is recovered using ISP and further refined.}
    \label{fig:pipeline}
\end{figure*}

\section{Related Work}
\label{sec:related}

Before the advent of Deep Learning, garment shape recovery from images depended mostly on user defined outlines and shape-from-shading techniques~\cite{Zhou13b}. Since then, data-driven techniques have become dominant.

\parag{Tight-Fitting Clothing.}

The majority of methods developed in recent years focus on clothing that clings relatively closely to the body.  These can be classified into two main categories. 

In the first category, are methods that model garments as surfaces that are distinct from the body surface and interact with it. DeepGarment \cite{Danerek17}, MGN \cite{Bhatnagar19}, and BCNet \cite{Jiang20d} train neural networks on synthetic images to predict the vertex positions of specific mesh templates. \cite{Casado22} and \cite{Liu23b} leverage normal estimation to optimize the vertex position of the template and recover wrinkle details. Such methods are inherently limited in the range of shapes they can handle, and those being trained on synthetic data can easily fail when facing real images. To overcome these limitations, SMPLicit \cite{Corona21}, DIG \cite{Li22c}, and ClothWild \cite{Moon22} leverage Signed Distance Functions (SDF) to recover a wide array of garment meshes from RGB images and the corresponding segmentation masks. However, to represent non-watertight garment surfaces using an SDF, one has to wrap around them a watertight surface with a minimum thickness, which reduces accuracy. This can be addressed by using Unsigned Distance Functions (UDFs) instead~ \cite{Guillard22b,DeLuigi23} but creates robustness issues: if the UDF is even slightly inaccurate, the value of the surface is never exactly zero and holes can appear in the reconstructed models. In our experience, the Implicit Sewing Patterns (ISP) model of~\cite{Li23a} effectively addresses the issues of generality, accuracy, and robustness. The garments consist of flat 2D panels whose boundary is defined by a 2D SDF. To each panel is associated a 3D surface parameterized by the 2D panel coordinates. Hence, different articles of clothing are represented in a standardized way, which allows the recovery of various garments from single images. This is why we choose it as the basis for our approach.

In the second category, are the many methods that represent body and garment using a single model. For example, in~\cite{Jackson18,Zheng19} a volumetric regression network yields a voxel representation of 3D clothed humans given a single image. Other works~\cite{Saito19a,Saito20a,He20d,Alldieck22} employ a pixel-aligned implicit function that defines 3D occupancy fields or signed distance fields for clothed humans. In~\cite{Alldieck19b,Alldieck19a},  displacement vectors or UV maps are used to represent deviations from a SMPL parametric body model~\cite{Loper15}. Similarly, in~\cite{Huang20c,He21,Xiu22,Zheng21} parametric body models are combined with implicit representations to achieve robustness to pose changes. While effective, all these methods suffer from significant limitations, because they cannot separate the surface of the garment from that of the body, and they are at a disadvantage when it comes to modeling loose garments whose motion can be relatively independent from the body.

\parag{Loose-Fitting Clothing.}
There is a more limited number of methods designed to handle free-flowing garments. Some recent works~\cite{Yang18f,Zhu20,Zhu22} rely on complex physics simulation steps or feature line estimation to align the surface reconstruction with the input image. However, their dependence on garment templates limits their generality, in the same way it did for other template-based methods discussed above.  Point-based methods that can reconstruct generic clothes have been proposed~\cite{Zakharkin21,Srivastava22} to overcome this. Unfortunately, point clouds are not straightforward to integrate into downstream applications. As a result, the method of~\cite{Srivastava22} resorts to modified Poisson Surface Reconstruction (PSR) to create a garment surface from the point cloud, which can result in incorrect geometry.  Another point-based representation is introduced in~\cite{Ma22a}. While being successful at modeling and animating humans wearing loose garments, it also relies on PSR to infer the mesh from the point cloud, yielding  a single mesh that represents body and garment jointly. Furthermore, \cite{Ma22a} does not explore how this representation can be fitted to images.

ECON~\cite{Xiu23} is a method specifically designed for clothed human recovery from images. By leveraging techniques such as normal integration and shape completion, it achieves visually appealing results for individuals wearing loose clothing. However, as~\cite{Ma22a}, ECON produces a single watertight mesh that tightly binds the body and garment together, precluding easy use for applications such as cloth simulation and re-animation.


\section{Method}
\label{sec:method}

Given an image of a clothed person and a body model extracted from it using existing techniques, our goal is to recover accurate 3D models of the garments matching the image. To this end, we add to the Implicit Sewing Pattern (ISP) garment model~\cite{Li23a}, which provides us with a shape prior for garment in its rest state, a deformation model that allows us to recover its potentially large deformations, as  illustrated by Fig. \ref{fig:pipeline}. As a result, whereas the original ISP, like most current clothes-recovery algorithms, is limited to tight-fitting clothing, our approach can handle both tight- and loose-fitting garments, such as skirts and open jackets. 

In this section, we first describe briefly the ISP model upon which we build our approach. We then introduce the deformation model that underpins the main contribution of this paper, going from tight fitting clothes to loosely fitting and free flowing ones. Finally, we present our approach to fitting this model to real-world images. 

\subsection{ISP Garment Model}
\label{sec:isp}


\begin{figure*}[!ht]
    \centering
    \includegraphics[width=0.9\textwidth]{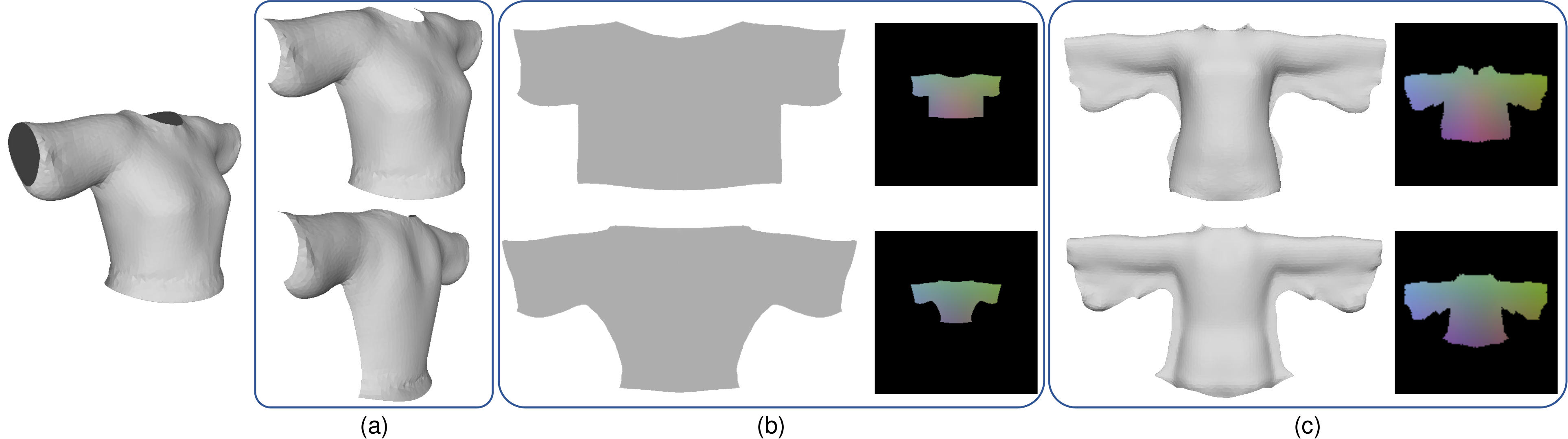}
    \caption{\textbf{Cutting and flattening.} (a) The front (top) and back (bottom) surfaces after cutting. (b) The flattened panels and UV maps generated by ISP for (a). (c) The maximum-coverage UV maps and its represented 3D shape.}
    \label{fig:flatten}
    \vspace{-0.2cm}
\end{figure*}

ISP is a garment model inspired by the sewing patterns that fashion designers use to represent clothes. A sewing pattern is made of several 2D panels along with information about how to stitch them into a complete garment. ISP implicitly models patterns using a 2D signed distance field and a 2D label field. 

\parag{Formalization.} 

Given the latent code $\bz$ of a garment and a point $\bu$ in the 2D UV space $\Omega=[-1,1]^2$ of a 2D panel of that garment, ISP outputs the signed distance $s$ to the panel boundary and a label $l$ as
\begin{equation} \label{eq:pattern}
    (s,l) = \mathcal{I}_{\Teta}(\bu,\bz) \; , 
\end{equation}
where $\mathcal{I}_{\Teta}$ is a fully connected network. The zero crossing of the SDF defines the shape of the panel, with $s<0$ indicating that $\bu$ is within the panel and $s>0$ indicating that $\bu$ is outside the panel. The stitch information is encoded in $l$, where panel boundaries with the same label should be stitched together. To transform the 2D sewing patterns into 3D surfaces, a UV parameterization function $\mathcal{A}_{\Phi}$ is learned to perform the 2D-to-3D mapping
\begin{equation} \label{eq:uv}
    \bX = \mathcal{A}_{\Phi}(\bu,\bz) \; ,
\end{equation}
where $\bX\in\mathbb{R}^3$ represents the 3D position of $\bu$. Essentially, ISP registers each garment onto a unified 2D space $\Omega$, and represents it using UV maps that record 2D-to-3D mapping, as shown in Fig. \ref{fig:flatten}(b). Given the paired 2D sewing patterns and their 3D meshes, the pattern parameterization network $\mathcal{I}_{\Teta}$ and the UV parameterization network $\mathcal{A}_{\Phi}$ are trained by minimizing the losses
\begin{align}
    L_{\mathcal{I}} &= L_{SDF} + L_{CE} + ||\bz||_2^2\;  \label{eq:loss_pattern} \;  , \\
    L_{\mathcal{A}} &= L_{MSE} + L_{consist} \; \label{eq:loss_uv}  \; , 
\end{align}
respectively. $L_{SDF}$ is the mean absolute error for the predicted SDF value $s$, $L_{CE}$ is the cross-entropy loss for the predicted label $l$, $L_{MSE}$ is the mean squared error of the predicted 3D position $\bX$, and $L_{consist}$ is the loss to reduce the gap between the front and back panels. More details can be found in \cite{Li23a}.

\parag{Training.}
Training ISP requires the 2D sewing patterns of 3D garments in a rest state, that is, draped on a T-pose neutral body. However, patterns are absent in most large scale garment datasets, including CLOTH3D \cite{Bertiche20}. We use the garment flattening algorithm of ~\cite{Pietroni22} to generate the required 2D patterns. The garment mesh is first divided into several pieces given predefined cutting rules. These pieces are then unfolded into 2D panels by minimizing an as-rigid-as-possible~\cite{Liu08c} energy, ensuring local area preservation between the 3D and 2D parameterizations. While doing this, we constrain the boundary vertices at places such as waist and sleeves to have a constant value along a specific axis. This enhances pattern consistency across the dataset. Fig. \ref{fig:flatten} illustrates this and shows the panels generated for a shirt. We provide more details in the supplementary material.  

We generate a front and a back panels as the sewing pattern for each garment in our dataset (shirt, skirt, and trousers). Once ISP has been trained on these, we compute the maximum-coverage UV maps $\bar{\mathcal{M}}$ over the UV maps of each garment for each category, as shown in Fig. \ref{fig:flatten}(c). The values of $\bar{\mathcal{M}}$ are taken to be
\begin{equation} \label{eq:m_bar}
    \bar{\mathcal{M}}[u,v] = \frac{\sum_i \mathbbm{1}_{s_{\bu}^i\le0} \cdot m_i[u,v]}{{\sum_i \mathbbm{1}_{s_{\bu}^i\le0}}},
\end{equation}
where $m_i$ is the UV maps of garment $i$, $\mathbbm{1}$ is the indicator function, $[\cdot, \cdot]$ denotes array addressing, and $s_\bu^i$ is the SDF value of ISP at $\bu=(u,v)$. The {\it maximum coverage map} $\bar{\mathcal{M}}$ encompasses information from all the patterns in the dataset. It represents the smallest possible map that covers all garments in a category, with the 3D position of each {\it uv}-point $\bu$ being the average of all garments that include $\bu$. We use it as a prototype for a garment category and to compute an initial guess of the deformed garment given the body pose, as discussed below.

\parag{Skinning.} 
As in many prior work~\cite{Li22c,DeLuigi23,Li23a}, we use SMPL~\cite{Loper15} to parameterize the body in terms of shape and pose parameters $(\beta,\theta)$, and its extended skinning procedure for the 3D volume around the body to initially deform the 3D shape represented as $\bar{\mathcal{M}}$. More specifically, given a 2D point $\bu=(u,v)$ in UV space $\Omega$, we get its actual 3D position as $\bar{\bX}_\bu=\bar{\mathcal{M}}[u,v]$. We then deform it by computing
\begin{align}
    \bX_\bu &= W(\bX_{(\beta,\theta)}, \beta, \theta, w(\bar{\bX}_\bu)\mathcal{W}) \;  \label{eq:smallDef} , \\
    \bX_{(\beta, \theta)} &= \bar{\bX}_\bu +  w(\bar{\bX}_\bu)B_\beta + w(\bar{\bX}_\bu)B_\theta \; , \nonumber
\end{align}
where $W(\cdot)$ is the SMPL skinning function with skinning weights $\mathcal{W} \in \mathbb{R}^{N_B\times 24}$, with $N_B$ being the number of vertices of the SMPL body mesh, and $B_{\beta} \in \mathbb{R}^{N_B\times 3}$ and $B_{\theta}\in \mathbb{R}^{N_B\times 3}$  are the shape and pose displacements of SMPL, respectively. The diffused weights $w(\cdot)\in \mathbb{R}^{N_B}$ are computed by a neural network, which generalizes the SMPL skinning to any point in 3D space. By repeating this for all points in a panel, we obtain the  vertex position map $\mathcal{M}[u,v]=\bX_\bu$.

\subsection{Modeling Large Deformations}
\label{sec:deform}

As shown in Fig. \ref{fig:pipeline}, the ISP approach described above generates a garment prototype $\mathcal{M}$ that closely fits the underlying body.  To a point $\bu = (u,v)$ that belongs to a given garment panel, it associates the vertex position $\bX_\bu=\mathcal{M}[u,v]$ of Eq.~\ref{eq:smallDef}, which is usually relatively close to the body. To model loose clothing, we now need to compute a potentially larger displacement $\Delta \bX_\bu$ to be added to $\bX_\bu$.

To evaluate $\Delta \bX_\bu$, we train a network $\mathcal{D}$ that consists of an MLP to estimate the occupancy value and the corrective displacement value, and two CNNs to extract image features $F_n$ and $F_b$ from the  image of normals and the segmentation and vertex position images of the SMPL body, respectively. Fig. \ref{fig:pipeline} depicts this architecture. We obtain the pixel-aligned image feature for $\bu$ by computing
\begin{align}
    F(\bx_{\bu}) &= F_n \oplus F_b(\bx_{\bu}) \;  \label{eq:feature} , \\
    \bx_{\bu} &= P(\bX_\bu) \; \label{eq:proj} , \nonumber
\end{align}
where $P(\cdot)$ denotes the projection into the image and $\oplus$ concatenation. The MLP takes as input $\bu$, its 3D position and image features to predict the occupancy $O_\bu\in\{0,1\}$ and the corrective displacement $\Delta \bX_\bu$. By assembling the results of each point in the UV space, we obtain the final occupancy maps $\mathcal{O}$ and the vertex position maps $\hat{\mathcal{M}}$, where $\hat{\mathcal{M}}[u,v]=\bX_\bu+\Delta \bX_\bu$. In essence, $\mathcal{O}$ is the binarized SDF of the garment 2D panels, which implicitly defines the garment shape and geometry in the rest state, while $\hat{\mathcal{M}}$ encodes the deformed state for that garment. The 3D garment mesh can be recovered from these 2D maps with ISP, as discussed below. 

We use a single network $\mathcal{D}$ for the front and the back panels, and encode the points $\bu$ of the front as $[u,v,+1]$ and those of the back as $[u,v,-1]$. We learn the parameters of $\mathcal{D}$ by minimizing
\begin{equation}\label{eq:loss_D}
    L = \sum_{\bu\in\Omega}||\bX_\bu+\Delta \bX_\bu-\tilde{\bX}_\bu||_2+ \lambda\sum_{\bu\in\Omega}BE(O_\bu, \tilde{O}_\bu),
\end{equation}
where $\tilde{\bX}_\bu$ and $\tilde{O}_\bu$ are the ground truth vertex positions and occupancy values, $BE(\cdot)$ is the binary cross entropy, and $\lambda$ is a weighting constant.

\parag{From 2D maps to 3D mesh.} 
The occupancy value $O_\bu$ indicates whether $\bu$ falls within the panels of the garment. To convert the generated 2D occupancy maps $\mathcal{O}$ into a 3D garment mesh in its rest shape, we need to recover an ISP latent code $\bz$ as defined in Eq.~\ref{eq:pattern}. To this end, we find the vector $\bz$ such that the corresponding SDF of ISP best matches the produced occupancy maps by minimizing
\begin{small}
\begin{equation} \label{eq:z} 
        \bz^* = \argmin\limits_{\bz} \sum\limits_{\scalebox{.6}{$\bu\in\Omega_-$}}R(s_\bu(\bz))+ \sum\limits_{\scalebox{.6}{$\bu\in\Omega_+$}}R(-s_\bu(\bz)) + \lambda_\bz||\bz||_2 \; ,
\end{equation}
\end{small}
where $\Omega_-=\{\bu|O_\bu=1, \bu\in\Omega\}$, $\Omega_+=\{\bu|O_\bu=0, \bu\in\Omega\}$, $R(\cdot)$ is the ReLU function, $s_\bu(\bz)$ is the SDF value of $\bu$ computed by ISP, and $\lambda_\bz$ is the weighting constant. With $\bz^*$, the rest garment mesh is inferred through ISP's meshing and sewing process. The deformed garment mesh can be obtained by simply replacing the vertex position of the recovered mesh with the values stored in $\hat{\mathcal{M}}$. This yields 3D garment meshes in both the rest and deformed states as shown in the top-right of Fig.~\ref{fig:pipeline}, which are required for the application such as cloth simulation and can be used for further refinement as discussed below.

\subsection{Fitting the Models to Images}
\label{sec:in_the_wild}


\begin{figure}[!ht]
    \centering
    \includegraphics[width=0.4\textwidth]{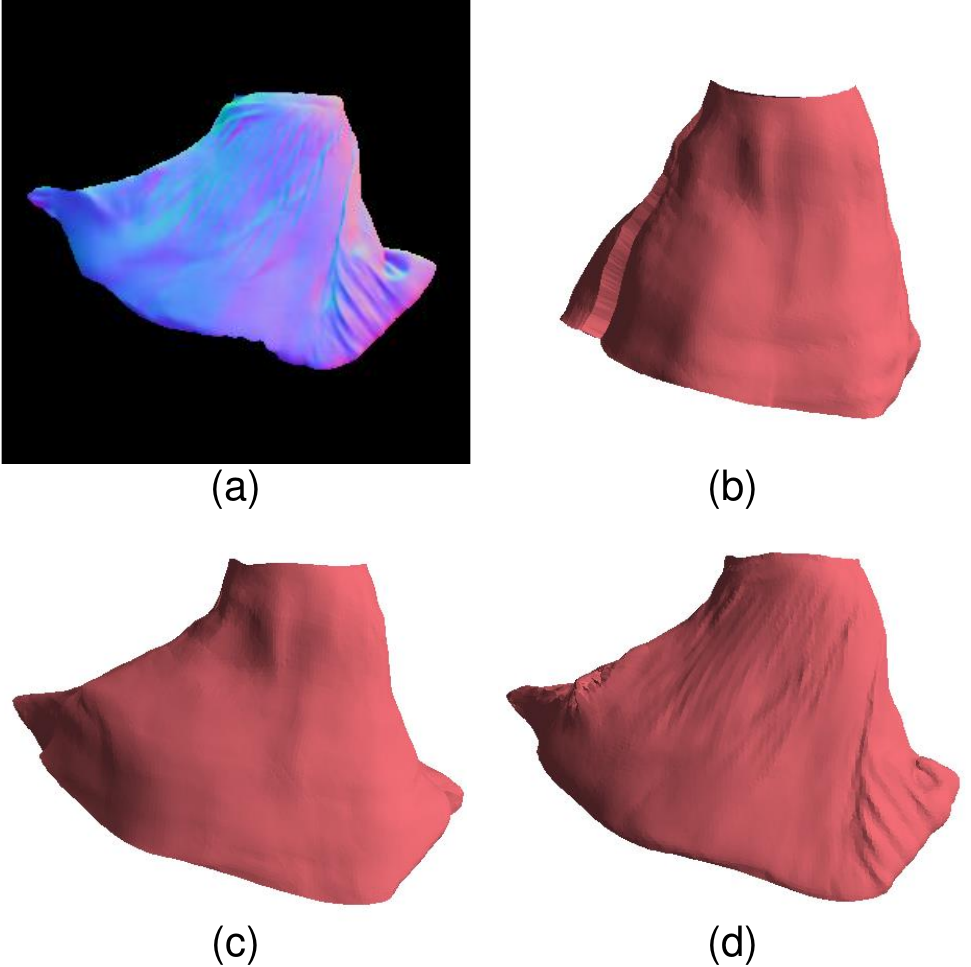}
    \caption{\textbf{Fitting results.} Given (a) the normal estimation of an in-the-wild image, (b) is the inference result with ISP recovered geometry. (c) is obtained by optimizing the parameters of the pre-trained deformation model. Further refinement of the mesh vertex positions yields (d).}
    \label{fig:opt}
    \vspace{-0.25cm}
\end{figure}

For practical reasons, the range of garment materials, external forces, and body motions present in the training data is limited. As a result, given in-the-wild images as input, the trained model can produce inaccurate results as shown in Fig. \ref{fig:opt}(b). To remedy this and to leverage the deformation prior that the network $\mathcal{D}$ captures, we refine the result by minimizing a loss function with respect to the pretrained deformation parameters of $\mathcal{D}$ as in~\cite{Ulyanov18,Joo20,Gadelha21}. $L$ is designed to promote a good match between the garment mesh and image observations. We take it to be
\begin{align}
    L &= \lambda_C L_{CD} + \lambda_n L_{normal} + \lambda_p L_{physics} \;  \label{eq:optimization} , \\
    L_{CD} &= d(x_c^f,x_{I_n}) \; \label{eq:cd} , \\
    L_{normal} &= \sum_{i\in f} 1 - cos(n^i, I_n(x_c^i)) \; \label{eq:normal} , \\
    L_{physics} &= L_{strain} + L_{bend} + L_{gravity} + L_{col}  \; \label{eq:physics} ,
\end{align}
where $x_c^f$ is the 2D projection of the centers of visible faces $f$ after mesh rasterization, $x_{I_n}$ denotes the coordinates of foreground pixels, $n^i$ is the normal of face $i$, $I_n(x_c^i)$ is the normal image values at $x_c^i$, and $\lambda_C$, $\lambda_n$ and $\lambda_p$ are the balancing scalars. $d(\cdot)$ and $cos(\cdot)$ are the functions measuring the 2D Chamfer Distance and the cosine similarity, respectively. $L_{physics}$ is a physics-based loss derived from \cite{Narain12,Santesteban22}, which computes the membrane strain energy $L_{strain}$ caused by the deformation, the bending energy $L_{bend}$ resulting from the folding of adjacent faces, the gravitational potential energy $L_{gravity}$ and the penalty for body-garment collision $L_{col}$. Minimizing $L_{CD}$ induces an external force of stretching or compression on the garment mesh to align its 2D projection with the given image, while minimizing $L_{physics}$ ensures that the mesh exhibits physically plausible deformation adhering to the shape constraints of the rest-state mesh recovered by ISP.

Minimizing the loss $L$ with respect to the deformation parameters yields a mesh whose overall shape matches the input image, as illustrated in Fig. \ref{fig:opt}(c). However, since neural networks tend to learn low-frequency functions \cite{Rahaman19}, the result might be too smooth. To recover fine surface details, we perform a refinement step by minimizing $L$ directly with respect to the coordinates of the garment mesh vertices. This generates realistic local details, such as wrinkles on the surface, as shown in Fig. \ref{fig:opt}(d).

\section{Experiments}
\label{sec:Experiments}

In short, our method begins by inferring the shape and the deformation of the garment in terms of the occupancy and position maps. Leveraging the shape prior of ISP, we then recover the garment geometry from the occupancy maps and deform it using the position maps. Next, we refine the initial deformed mesh to better align with image observations by fine-tuning the pre-trained network $\mathcal{D}$ that captures the deformation prior. Finally, we recover fine details through vertex-level optimization of the garment mesh. In this section, we demonstrate the effectiveness of this process and compare it to that of other state-of-the-art methods. 

\subsection{Implementation Details}

Following the implementation of \cite{Li23a}, both the pattern parameterization model $\mathcal{I}_\Theta$ and the  UV parameterization model $\mathcal{A}_\Phi$ of ISP have two separate MLPs for the front and back panels. Each MLP has 7 layers with Softplus activations. The dimension of latent code $\bz$ is 32. The skinning weight model $w$ is a 9-layer MLP with leaky ReLU activations, whose output is normalized by a final Softmax layer. $\mathcal{I}_\Theta$ and $\mathcal{A}_\Phi$ are trained jointly for 9000 iterations with a batch size of 50. $w$ is trained with the same parameters as \cite{Li23a}. For the image feature extraction, we use two separate ConvNeXt \cite{Liu22e} networks to extract multi-scale garment and body features of sizes 96, 192, 384, 768, which are concatenated as the final features. The point UV coordinates $\bu$, 3D position $\bX_\bu$, and image features $F(\bx_\bu)$ are projected separately to 384 dimensions by three linear layers, which are then concatenated as the input of the MLP of the deformation model. The MLP of the deformation model has 10 layers with a skip connection from the input layer to the middle, and uses Gaussian functions as the activation layer following \cite{Ramasinghe22}. The CNNs and MLP of the deformation model are trained jointly for 40 epochs with the Adam optimizer \cite{Kingma15a} and a learning rate of $10^{-4}$.  For real images, we use \cite{Xiu23} and \cite{Feng21} to obtain their normal and SMPL body parameter estimations, respectively. The garment segmentation masks are generated by leveraging the segmentation of SAM \cite{Kirillov23} and the semantic labels of \cite{Li20i}.


\begin{figure*}[!ht]
    \centering
    \includegraphics[width=0.99\textwidth]{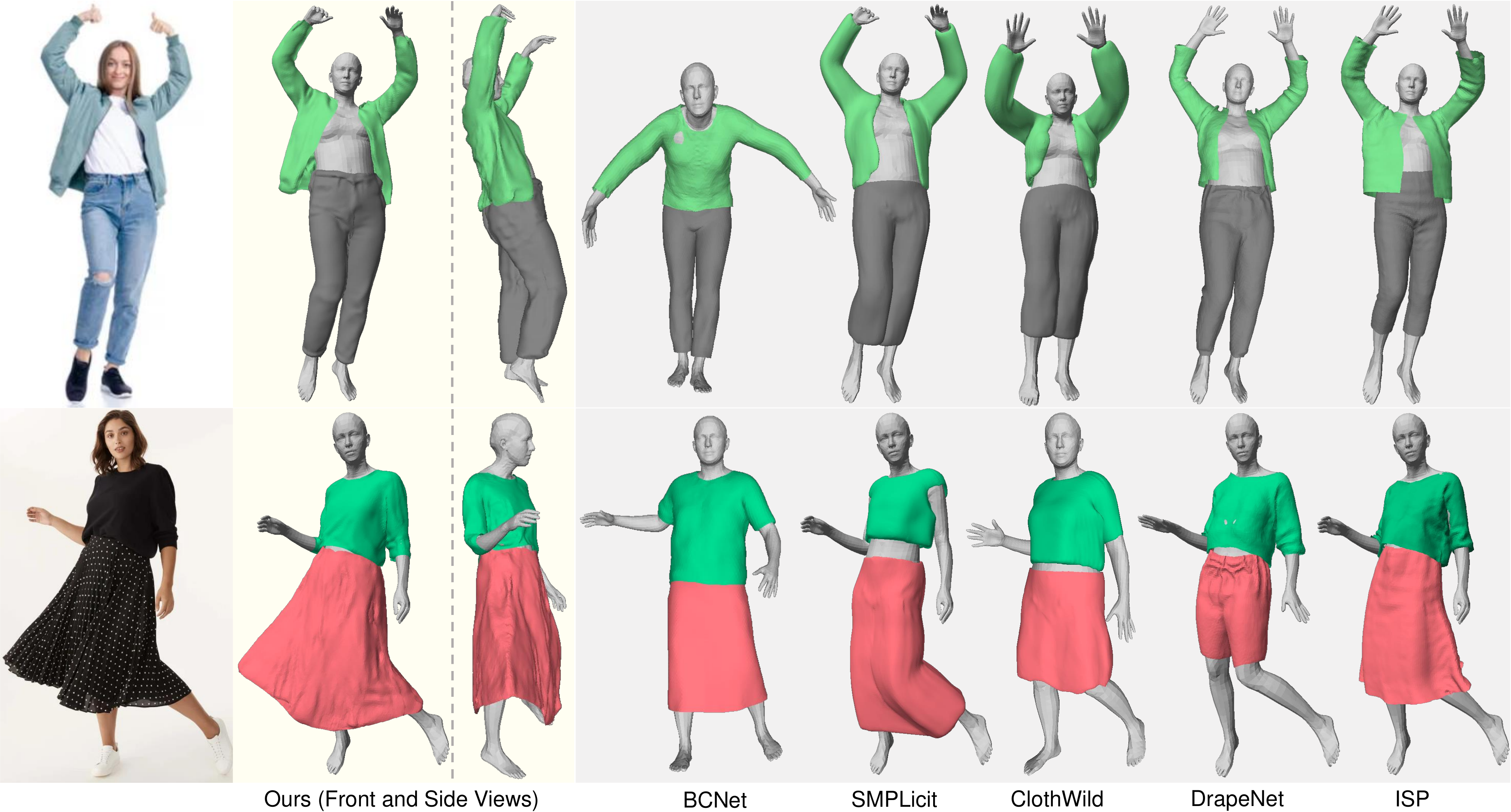}
    \vspace{-2mm}
    \caption{\textbf{Comparison against SOTA methods.} From left to right, we show the input image and the 3D garment meshes recovered by our method and SOTA methods: BCNet, SMPLicit, ClothWild, DrapeNet, ISP. (Since skirt is unavailable for DrapeNet, a random pair of trousers is put on its result in the second row.)}
    \label{fig:compare}
    \vspace{-0.2cm}
\end{figure*}

\subsection{Dataset, Evaluation Metrics, and Baseline}
Our models are trained on CLOTH3D \cite{Bertiche20}, which is a synthetic dataset with motion sequences of 3D clothed human. It contains garment in the rest state, and deformed states caused by the motion of underlying body. For each garment, it has a single simulated sequence up to 10 seconds. It covers a large variety of garment in different shapes, types and topologies. For the training of ISP, we randomly select 400 shirts, 200 skirts and 200 pairs of trousers, and generate their sewing patterns by the method described in Sec. \ref{sec:isp}. The deformation model is trained on the corresponding simulated sequences. For each frame, we render 11 normal images for the garment mesh with random rotations around the Y-axis, which produces 40K, 40K and 20K training images for shirt, skirt and trousers respectively. During training, we augment the data with image flipping and rotation. 

To evaluate the garment reconstruction quality, we use the Chamfer Distance (CD) between the ground truth and the recovered garment mesh, and Intersection over Union (IoU) between the ground truth mask and the rendered mask of reconstructed garment mesh.

We compare our method against state-of-the-art methods BCNet \cite{Jiang20d}, SMPLicit \cite{Corona21}, ClothWild \cite{Moon22}, DrapeNet \cite{DeLuigi23} and ISP \cite{Li23a}. SMPLicit, DrapeNet and ISP use the garment segmentation mask for reconstruction, while BCNet and ClothWild take the RGB images as input. 

\subsection{Comparison with State-of-the-Art Methods}


\begin{table}[ht!]
    \begin{center}
    \begin{tabular}{cc}
    \hspace{-5mm}
      \scalebox{0.7}{
            \begin{tabular}{c | c | c | c}
            \toprule
              CD ($\times10^3$) $\downarrow$& Skirt & Shirt & Trousers \\
             \midrule
             SMPLicit & 8.05 & 3.49 & 0.81 \\
             DrapeNet & n/a  & 1.54 & 0.84 \\
             ISP      & 7.26 & 2.02 & 0.91\\
             \midrule
             Ours     & \textbf{2.87} & \textbf{0.44} & \textbf{0.28}  \\
              \midrule  
             \textit{Ours-GT} & \textit{1.67}  & \textit{0.31} & \textit{0.23} \\
            \bottomrule
            \end{tabular}
            }
        &
         \hspace{-5mm}
        \scalebox{0.7}{
            \begin{tabular}{c | c | c | c}
            \toprule
              IoU $\uparrow$ & Skirt & Shirt & Trousers \\
             \midrule
             SMPLicit & 0.657 & 0.527 & 0.782 \\
             DrapeNet & n/a  & 0.715 & 0.845 \\
             ISP      & 0.709 & 0.665 & 0.781\\
                \midrule
             Ours     & \textbf{0.940} & \textbf{0.939} & \textbf{0.946} \\
             \midrule
             \textit{Ours-GT} & \textit{0.953}  & \textit{0.942} & \textit{0.954} \\
            \bottomrule
            \end{tabular}
            }  
      \end{tabular}
      \end{center}
      \vspace{-5mm}
      \caption{\textbf{Quantitative comparisons.} Our method outperforms SMPLicit, DrapeNet, and ISP in terms of CD and IoU on all three garment categories, as shown in the second-to-last row of both tables (Ours). These results were obtained using normals estimated from the images using~\cite{Xiu23}. In the last row (Ours-GT), we provide the results we obtained using ground-truth normals instead.}
      \label{tab:synthetic}
  \end{table}

\parag{Quantitative Results.}
Due to the absence of publicly available real dataset for evaluating garment reconstruction, we utilize synthetic data consisting of 30 unseen shirts, 30 unseen skirts, and 30 pairs of unseen trousers from CLOTH3D for quantitative evaluation. We compare our method with SMPLicit, DrapeNet, and ISP, which recover garment meshes from segmentation masks. The segmentation masks required by the baselines are rendered from the ground truth mesh data, while the normal images required by our method are estimated with~\cite{Xiu23}. As shown in Tab.~\ref{tab:synthetic}, our approach significantly outperforms the baselines in terms of Chamfer Distance (CD) and Intersection over Union (IoU) across all garment categories.

For comparison purposes, we re-ran our algorithm using the ground-truth normals and report the results in the last row of Tab.~\ref{tab:synthetic}. As expected, this leads to an improvement in reconstruction accuracy, but the increase is only modest. This highlights the robustness of our approach to the slight inaccuracies that can be expected from a normal-estimation algorithm.

\parag{Qualitative Results.}
Fig. \ref{fig:compare} shows the qualitative comparison for the results reconstructed from in-the-wild images. Since BCNet is trained only on synthetic RGB data, it is not able to predict accurate body and garment results. Both SMPLicit and ClothWild generate unrealistic watertight meshes that do not accurately represent real garments. While DrapeNet and ISP are able to recover open surfaces for garments, their results remain closely adhered to the body, similar to SMPLicit and ClothWild. In contrast, our method can faithfully recover garment mesh from input images, as clearly observed in the depiction of the open jacket and the flowing long skirt that stand away from the body.


\begin{table}
    \begin{center}
      \scalebox{.89}{
            \begin{tabular}{c | c | c }
            \toprule
              CD ($\times10^3$) $\downarrow$ & GT & Est. \\
             \midrule
             (-$\mathcal{D}^*$,-$v^*$)  & 2.04 & 3.58  \\
             (-$\mathcal{D}^*$,+$v^*$) & 1.80  & 3.39  \\
             (+$\mathcal{D}^*$,-$v^*$) & 1.75  & 2.93  \\
             \midrule
             (+$\mathcal{D}^*$,+$v^*$) & \textbf{1.67}  & \textbf{2.87}  \\
            \bottomrule
            \end{tabular}
            }
      ~
      \scalebox{.89}{
            \begin{tabular}{c | c | c }
            \toprule
              IoU $\uparrow$ & GT & Est. \\
             \midrule
             (-$\mathcal{D}^*$,-$v^*$)  & 0.848 & 0.751  \\
             (-$\mathcal{D}^*$,+$v^*$) & 0.947  & 0.904  \\
             (+$\mathcal{D}^*$,-$v^*$) & 0.941  & 0.923  \\
             \midrule
             (+$\mathcal{D}^*$,+$v^*$) & \textbf{0.953}  & \textbf{0.940}  \\
            \bottomrule
            \end{tabular}
            }
      \end{center}
      \caption{\textbf{Ablation study.} GT and Est. mean using ground-truth and estimated normal images as input, respectively. +/-$\mathcal{D}^*$ denote with/without the finetuning of the deformation model. +/-$v^*$ denote with/without the optimization of vertex positions.
      }
      \label{tab:ablation}
  \end{table}

\begin{figure}[!ht]
    \centering
    \includegraphics[width=0.47\textwidth]{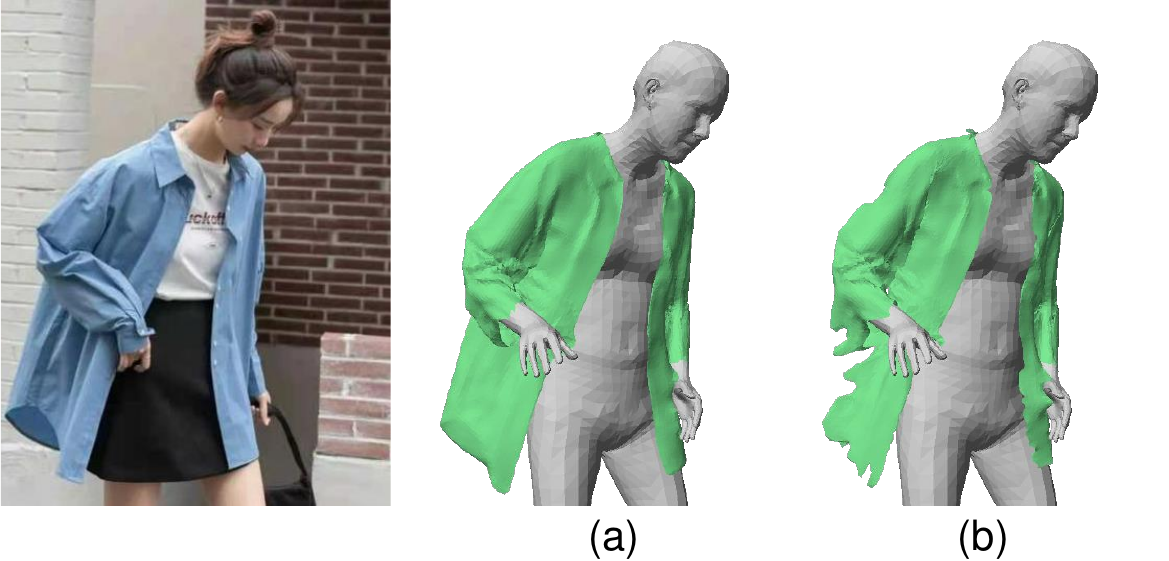}
    \caption{\textbf{Fitting strategy} comparison between  (a) our fitting method and (b) only vertex optimization.}
    \label{fig:ablation}
\end{figure}


\begin{figure*}[h]
    \centering
    \includegraphics[width=1.\textwidth]{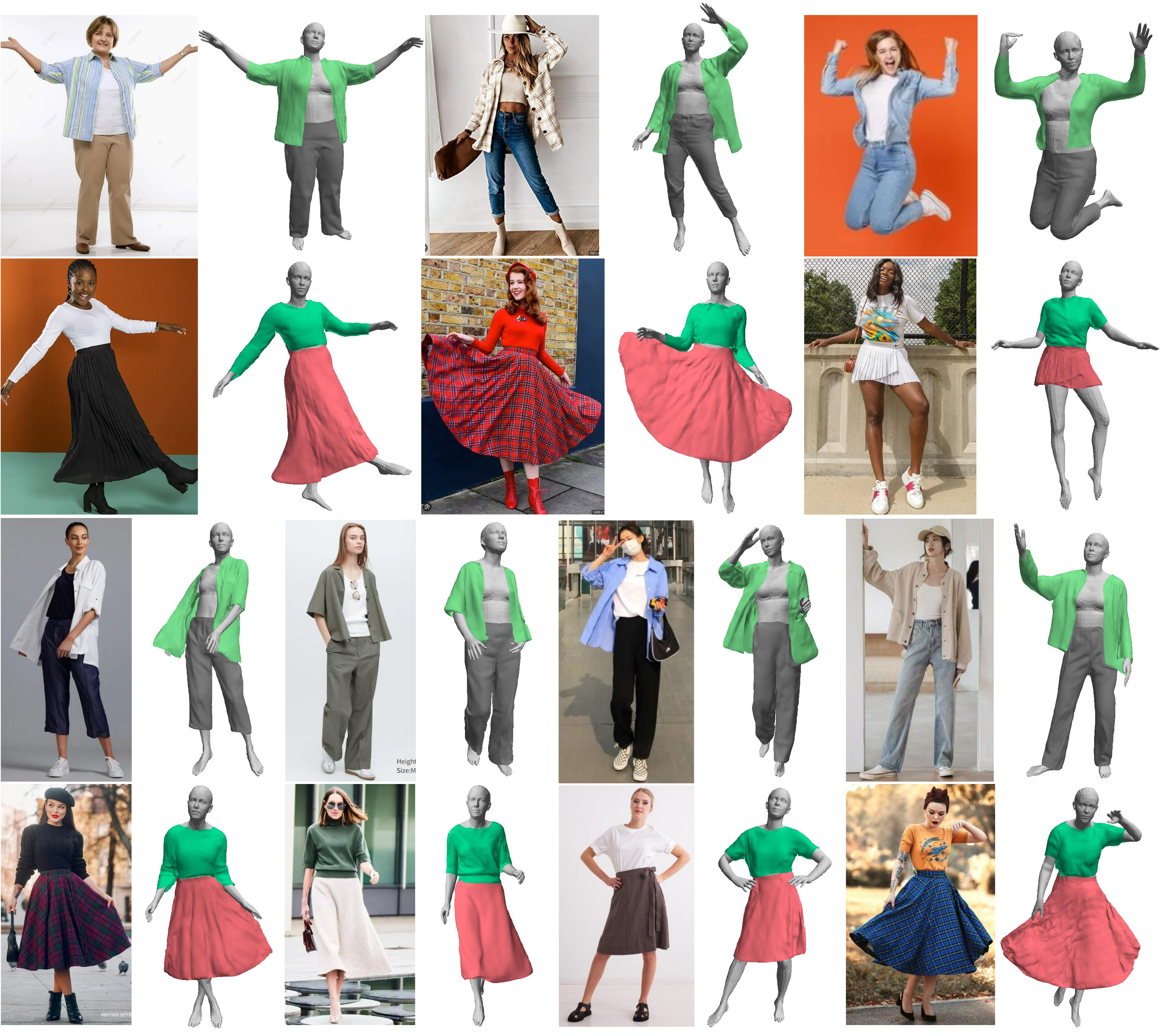}
    \caption{\textbf{Recovering from in-the-wild images.} Our method is able to recover realistic meshes for garments with diverse shapes and deformations.}
    \label{fig:results}
    \vspace{-0.25cm}
\end{figure*}

\subsection{Ablation Study}
Table \ref{tab:ablation} presents the evaluation results of our method with and without fitting on the test set of skirts. The results of row 1, 3 and 4 indicate that optimizing the parameters of the pretrained deformation network (+$\mathcal{D}^*$,-$v^*$) improves the quality of the raw inference results, and optimizing the vertex positions (+$\mathcal{D}^*$,+$v^*$) further enhances the outcome, yielding the lowest CD and highest IoU. 

However, as demonstrated in the second row, directly optimizing the vertex positions on the raw inference results without optimizing the deformation model (-$\mathcal{D}^*$,+$v^*$) is not as effective. The deformation model encapsulates a continuous deformation field. When its weights are optimized based on partial observations, the entire field undergoes modification, thereby influencing all mesh vertices globally. On the contrary, a direct vertex optimization with partial observations predominantly affects the mesh vertices locally. While this can capture localized details like wrinkles, it struggles to resolve discrepancies in the overall shape. Fig. \ref{fig:ablation} displays the results of our methods and an ablation with vertex optimization only. Unsurprisingly, the latter fails at recovering the correct shape and produces implausible deformation on the mesh surface.

\subsection{More Results}
Fig. \ref{fig:results} shows a collection of results reconstructed by our method from in-the-wild images. Our method can produce realistic 3D meshes with fine details across a wide range of garment types, from tight-fitting attire to more relaxed and flowing outfits.

\section{Conclusion}
\label{sec:conclusion}

We have presented a novel approach to recovering realistic 3D garment meshes from in-the-wild images featuring loose fitting clothing. It relies on a fitting process that imposes shape and deformation priors learned on synthetic data to accurately capture garment shape and deformations.  In future work, we will extend our approach to modeling deformations over time from video sequences while enforcing temporal consistency of the reconstructions.

\parag{Acknowledgement.} This project was supported in part by the Swiss National Science Foundation.

\clearpage
{\small
\bibliographystyle{ieeenat_fullname}
\bibliography{string,geom,graphics,learning,vision,misc,new_ref}
}


\end{document}